\documentclass[a4paper, 10pt, conference]{IEEEtran}
\IEEEoverridecommandlockouts
\usepackage{cite}
\usepackage{url}
\usepackage{amsmath,amssymb,amsfonts}
\usepackage{algorithmic}
\usepackage{graphicx}
\usepackage{textcomp}
\usepackage{xcolor}
\def\BibTeX{{\rm B\kern-.05em{\sc i\kern-.025em b}\kern-.08em
    T\kern-.1667em\lower.7ex\hbox{E}\kern-.125emX}}

\usepackage{lipsum}

\newcommand\blfootnote[1]{%
  \begingroup
  \renewcommand\thefootnote{}\footnote{#1}%
  \addtocounter{footnote}{-1}%
  \endgroup
}

\newcommand\T{\rule{0pt}{2.4ex}}       
\newcommand\B{\rule[-1.2ex]{0pt}{0pt}} 

\begin{document}

\title{A Feasibility Study on Image Inpainting for Non- cleft Lip Generation from Patients with Cleft Lip}
\author{
\IEEEauthorblockN{Shuang Chen$^{1}$, Amir Atapour-Abarghouei$^{1}$, Jane Kerby$^{2}$, Edmond S. L. Ho$^{3}$, \\David C. G. Sainsbury$^{2}$, Sophie Butterworth$^{2}$, Hubert P. H. Shum$^{1,\dag}$}
\IEEEauthorblockA{\small $^1$ Department of Computer Sciences, Durham University, Durham, United Kingdom}
\IEEEauthorblockA{\small $^2$ Department of Plastic and Reconstructive Surgery, The Newcastle Upon Tyne Hospitals NHS Foundation Trust}
\IEEEauthorblockA{\small $^3$ School of Computing Science, University of Glasgow, Glasgow, United Kingdom}
\vspace{-2.0em}
}


\maketitle

\begin{abstract}
A Cleft lip is a congenital abnormality requiring surgical repair by a specialist.
The surgeon must have extensive experience and theoretical knowledge to perform surgery, and Artificial Intelligence (AI) method has been proposed to guide surgeons in improving surgical outcomes. 
If AI can be used to predict what a repaired cleft lip would look like, surgeons could use it as an adjunct to adjust their surgical technique and improve results. 
To explore the feasibility of this idea while protecting patient privacy, 
we propose a deep learning-based image inpainting method that is capable of covering a cleft lip and generating a lip and nose without a celft.
Our experiments are conducted on two real-world cleft lip datasets and are assessed by expert cleft lip surgeons to demonstrate the feasibility of the proposed method.

\end{abstract}

\begin{IEEEkeywords}
Cleft Lip, Image Inpainting, Deep Learning, Multi-task
\vspace{-1.6em}
\end{IEEEkeywords}

\blfootnote{Emails: \{shuang.chen, amir.atapour-abarghouei\}@durham.ac.uk, jane.kerby@nhs.net, shu-lim.ho@glasgow.ac.uk, david.sainsbury@nhs.net, sophie.butterworth2@nhs.net, hubert.shum@durham.ac.uk}
\blfootnote{This work involved human subjects or animals in its research. Approval of all ethical and experimental procedures and protocols was granted by the host organisation, the Research Ethics Committee (REC), the Health Research Authority (HRA), and Health and Care Research Wales (HCRW), under Approval Nos. 19/LO/1690 and under IRAS Project ID:
240451.}
\blfootnote{$^\dag$ Corresponding author}

\section{Introduction}
A cleft lip is a congenital condition which arises during pregnancy in the early stages of development where the upper lip does not fuse together.
One in every 700 births in the UK has a cleft lip and palate \cite{cleft_lip_NHS}.
Patients with an orofacial cleft  require surgical treatment by a cleft lip and palate surgeon at an average age of three months to correct cleft lip \cite{wellens2006keys}. 


Achieving symmetry and improving nasolabial appearance is a fundamental goal of cleft lip surgery \cite{mosmuller2017development}. 
There are various surgical approaches to repairing a cleft lip. The most commonly used worldwide at present are a Millard repair or a Fisher repair \cite{patel2019comparison}. Repairing cleft lips is a specialist skill and training in the UK requires an extended period of subspeciality training. Evaluating the outcome of cleft lip and palate surgery is an essential part of being able to improve surgical technique.
The current gold standard for assessing outcomes is the Asher-McDade rating scale by using a 5-point standard scale to assess nasolabial profile, nasal symmetry, nasal form, and vermilion border \cite{asher1991development}.

Recently, with the rapid advancement of AI, technologies based on deep learning have emerged to locate cleft lip surgical annotation and incisions to facilitate surgery \cite{li2019clpnet}. This may help junior surgeons in the early stages of their career and also surgeons who may not be as familiar with repairing cleft lips. Other clinical applications include being able to predict the outcome of a cleft lip repair which would enable surgeons to adjust their surgical procedure to provide the best outcome possible.


Our work aims to generate a non-cleft lip from an image of a baby with a cleft lip while protecting patient privacy. 
StyleGAN \cite{karras2021alias} would be one candidate,
given its impressive performance in style transfer tasks. However, due to their confidential nature, images of an individual with a cleft lip are not readily available, 
making it challenging to feed the data-hungry StyleGAN.
Even with sufficient data for training, once approaches such as StyleGAN fall victim to a model inversion attack, malicious users will be able to steal photos of real patients with a cleft lip \cite{zhu2019deep}, posing a grave privacy risk to patients. In this case, image inpainting is preferable, since we do not use images of individuals with cleft lip for training, the patient photographs leave no traces in the model.

\begin{figure*}[ht]
\centerline{\includegraphics[scale=0.54]{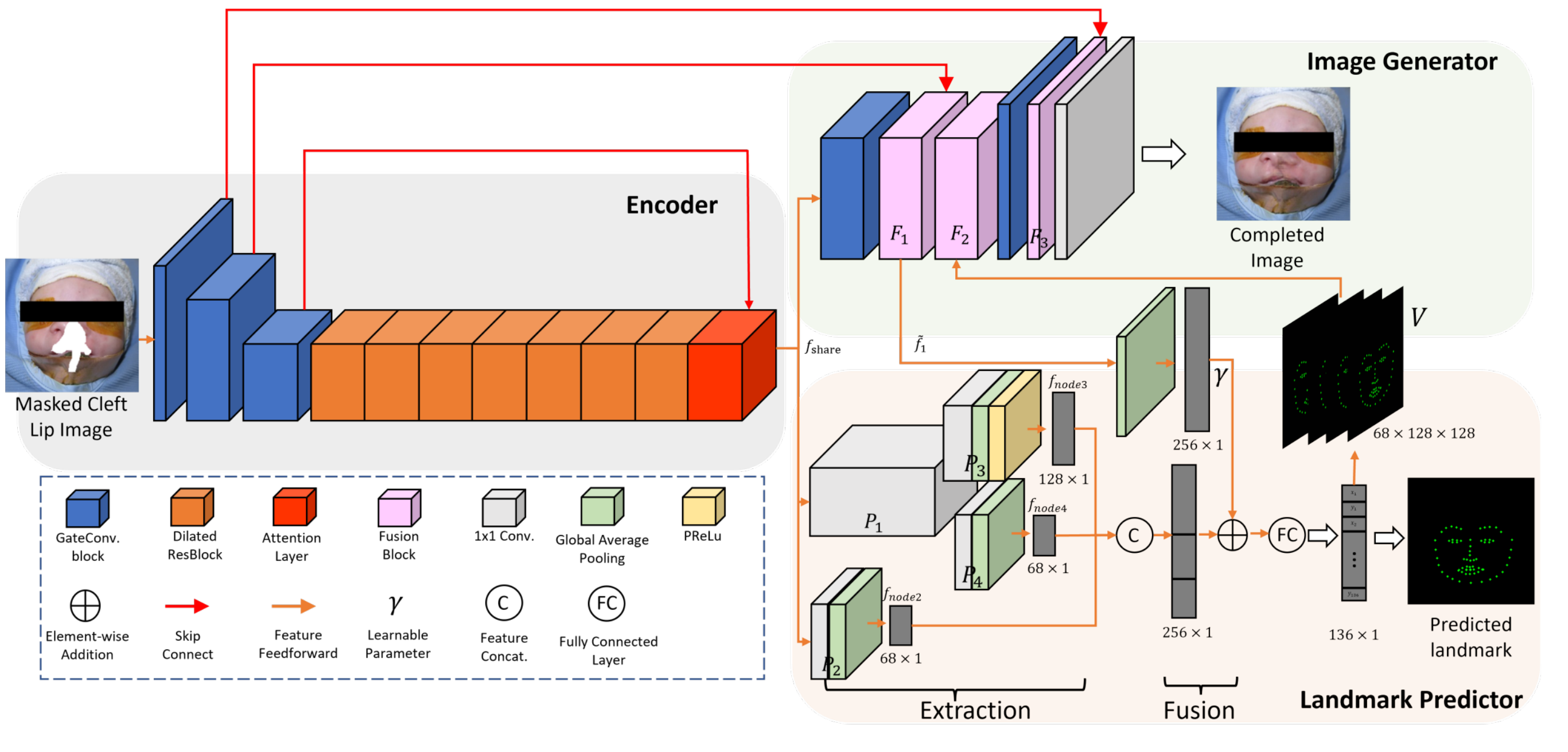}}
\vspace{-1.6em}
\caption{Overview of the proposed method.}
\label{framework}
\vspace{-1.6em}
\end{figure*}

In the facial inpainting task, the surgical evaluation criteria correspond to the semantic plausibility of the face shape and the image quality.
Existing advances in facial inpainting typically ensure the accuracy of generated facial attributes by supplying supplementary facial geometry information. 
EdgeConnect \cite{Nazeri_2019_ICCV} first generates a structure map, then combines the corrupted image to perform image inpainting in the second stage.
However, the correlation between structural information and texture is frequently redundant and unreliable \cite{liao2021image}. 
Human face can be modelled using landmarks and their geometrical features \cite{shi2006effective}. Lafin \cite{yang2019lafin} uses landmarks as indicators to more precisely described facial attributes. Nonetheless, both of EdgeConnect and Lafin have a multi-stage limitation: the final image quality is highly dependent on how well the indicator generation in the first stage works.

In this paper, we propose a single-stage end-to-end multi-task image inpainting framework to generate non-cleft lip from patients with cleft lip. 
We use adaptive feature fusion and landmark indicator to boost parameter sharing and utilize the second task more efficiently. The landmark prediction is guided by both masked image and partial inpainted information, resulting in a more precise geometry indicator for repairing facial attributes. Moreover, since we do not use images of individuals with cleft lips to train the model, the patient is protected against privacy violations.

To evaluate our model and inform the feasibility of our proposed method, we collected two datasets, CleftLip10 and CleftLip24, from real patients for testing. Specifically, we use segmentation masks to cover the cleft lip and medical equipment. Then our model automatically synthesizes a lip and nose without a cleft. We compare our results with EdgeConnect \cite{Nazeri_2019_ICCV}, Lafin \cite{yang2019lafin}, and CTSDG \cite{guo2021image}, which also challenged facial inpainting. We invited three professional cleft lip surgeons to rank the results. We also include quantitative results on CelebA \cite{liu2015deep} to show the advantages of our design.


Our main contributions are summarized as follows:
\begin{itemize}
\item We propose an image inpainting approach to produce an non-cleft lip image from patients with cleft lip. The code is publicly available to enable better reproducibility$^\star$.

\item We propose a multi-task network in which branches cooperate with each other through parameter sharing between tasks, which can achieve both landmark prediction and image inpainting at the same time. \end{itemize}
\blfootnote{$^\star$\url{https://github.com/ChrisChen1023/NCLG-MT}}

\section{Methodology}
We propose an end-to-end multi-task model, the framework is shown in Fig.~\ref{framework}. We first train our model on CelebA, then perform inference step on real patient images with cleft lip to generate non-cleft lip with semantic plausible facial attributes.
Our model can simultaneously perform image inpainting and facial landmark prediction. The paramters in two tasks are shared through
image-to-landmark and landmark-to-image feature fusion operations. Formally, the whole pipeline could be denoted as:

\begin{equation}
(\hat{I}, \hat{L})=G({I \odot(1-M})),
\label{pipeline}
\end{equation}
where $G$ is our multi-task model, $I$ is the real image and $M$ denotes the segmentation mask that occludes the cleft lip and medical equipment. $\hat{I}$ and $\hat{L}$ are completed image and predicted landmarks respectively.

\subsection{The Dataset}
To verify that our model can work on real clinical patient pictures, we collected CleftLip10 and CleftLip24 datasets. 
The datasets composed of frontal face images of patients who underwent a cleft lip repair at the Royal Victoria Infirmary (RVI) in Newcastle upon Tyne during the outpatient clinic. CleftLip10 contains images taken from 10 patients pre-operatively and immediately post-operatively. The images were taken using a Canon PowerShot G1 X Mark II Camera with a resolution of $3072 \times 2048$. CleftLip24 is comprised solely of pre-surgery images of twenty-four patients. The images were taken using a Canon EOS 20D or 5D Mark II DSLR Camera (with a 105mm lens) with a resolution of $2574 \times 3861$. All images were used in our experiments.

\subsection {Encoder and Image Generator} The encoder and the image generator jointly perform the inpainting task. 
The masked cleft lip image is downsampled three times and fed into the dilated convolutional residual blocks used to improve the receptive field, followed by a short-long attention layer to match feature more efficiently. We use gated convolutions instead of vanilla convolutions only in the image downsampling and downsampling stages. This is because 1) using gated convolution is more efficient for irregular masks \cite{yu2019free}, 2) its sensitivity to valid and missing pixels seems to be significant only for encoder and decoder \cite{cao2021learning} and 3) extensive use of gated convolution lead to a significant increase in parameter count. The shared feature is extracted at the end of the encoder:

\begin{equation}
f_{\text {share}}= E(I \odot(1-M),
\label{encoder}
\end{equation}
where $E$ is the encoder and $f_{\text{share}}$ is the deep feature from the attention layer (See Fig.\ref{framework} (Encoder)).

The image generator is designed to up-sample $f_{\text{share}}$ and reconstruct a non-cleft lip and nose. We employ three feature fusion blocks to facilitate parameter sharing, which are denoted by $F_{1}$, $F_{2}$, $F_{3}$ respectively. $F_{1}$ and $F_{3}$ aim to fuse the uncompleted image features from encoder by skip connections to generate more exquisite results by combining low-level and high-level feature.  


\begin{equation}
\widetilde{f}_{l}=\left\{\begin{array}{lr}
F_{i}\left(\operatorname{Concat}\left(f_{e i}, f_{d i}\right)\right), \text { if }(i=1,3) \\
F_{i}\left(\operatorname{Concat}\left(\widetilde{f}_{1}, V\right)\right), \text { if }(i=2)
\end{array}\right.,
\label{fusion}
\end{equation}
where $\widetilde{f}_{i}$ is the result from fusion block ${F}_i$ ($i=1,2,3$). ${f}_{e i}$ and ${f}_{d i}$ is the feature map from corresponding encoder and decoder layer. After $F_3$ followed by a vanilla convolution layer, completed image is genereated. $F_{2}$ is designed to fuse the Landmark map $V$ from the landmark predictor, which will be detailed in the next subsection.


\subsection{Landmark Predictor}
The landmark predictor involves extraction, fusion block and a fully-connected layer, aims to predict facial landmarks and inform the generator for assisting image inpainting.


The extraction step is designed to collect the landmark information from the encoded image feature. Specifically, $f_{\text{share}}$ is fed to a $1 \times 1$ convolutional layer $P_{1}$ to increase dimensionality, then we conduct dimensionality reduction followed by global average pooling to extract the feature into two vectors with different lengths ($P_{4}$ and $P_{3}$). Particularly, there is a $PReLu$ layer at the end of $P_{3}$ for non-linear projection. Simultaneously, $P_{2}$ also returns a vector after dimensionality reduction and global pooling directly acting on $f_{\text{share}}$, then we concatenate them:
\begin{equation}
f_{l m k}=\operatorname{Concat}\left(f_{\text {node2 }}, f_{\text {node3 }}, f_{\text {node4 }}\right),
\label{vector}
\end{equation}
where $f_{\text {nodei}}$ is the corresponding vector from $P_{i}$.
In existing multi-stage networks \cite{Nazeri_2019_ICCV, yang2019lafin}, generated indicators are assumed as perfect
and are used in final inpainting stage directly. A faulty indicator may mislead image inpainting.
To involve both corrupted and regenerated information in landmark predictor, and strengthen the parameter sharing between two tasks, we adaptively borrow $f_{1}$ from inpainting task,
followed by a global average pooling, we merge it with the concatenated landmark feature vector:
\begin{equation}
f_{l m k}^{\prime}=\gamma * \widetilde{f}_{1} \oplus f_{l m k},
\label{adaptive}
\end{equation}
where $\gamma$ is a trainable weight with zero initialization and $\oplus$ is element-wise addition. Finally, we apply a fully-connected layer to predict facial landmark points. 

To strengthen the parameter interaction between the two tasks and improve the completed image quality, we further map the landmark points into a binary feature map $V$, which is integrated with texture information in $F_{2}$. Formally, let $v_{pq}$ be the value in $V$ at position $(p,q)$: 

\begin{equation}
v_{p q}=\left\{\begin{array}{lr}
1, \text { if }\left(p=\left[\alpha x_{i}\right], q=\left[\alpha y_{i}\right]\right.) \\
0,  \text { otherwise }
\end{array} \right.,
\end{equation}
where $\alpha$ is a scale factor corresponding the size of the feature map in $F_{2}$, $[.]$ means integer operation. 
We create a $68 \times 128\times128$ tensor with landmark annotations, and transfer it to $F_2$ to provide facial geometry indicators. 
\subsection{Loss Function}
We follow Yang et al. (2019) to design our loss function, since our work applies a multi-task architecture, the landmark loss is involved in a joint loss function: 
\begin{equation}
\mathcal{L}_{l m k}=\left\|\hat{L}-L_{g t}\right\|_{2}^{2}.\label{LMK_Loss}
\end{equation}
We consider $L_{1}$ loss, adversarial loss, style loss, perceptual loss, total variation loss and landmark loss. Given a masked image $I$, the ground truth image $I_{gt}$ and corresponding landmark ground truth $L_{gt}$. The overall loss function is:
\begin{equation}
\begin{aligned}
\mathcal{L}_{\text {total}}(I, I_{gt},L_{gt})=\mathcal{L}_{\text {pixel }} &+\lambda_{\text {perc }} \mathcal{L}_{\text {perc }}+\lambda_{s t y} \mathcal{L}_{\text {style }} \\
&+\lambda_{t v} \mathcal{L}_{t v}+\lambda_{a d v} \mathcal{L}_{a d v_{G}}\\
&+\lambda_{l m k} \mathcal{L}_{l m k},
\end{aligned},\label{Total_Loss}
\end{equation}
where $\lambda_{\text {perc}}=\lambda_{sty}=\lambda_{tv}=0.1$, $\lambda_{adv}= 0.01$, $\lambda_{lmk}=0.00046$.

\section{Experiments}
\subsection{Training Details}

We train our model with CelebA \cite{liu2015deep}, which is a popular human face dataset containing over 160 thousands training face images. For CelebA, we remove a few images which can not be obtained landmark ground truth.
During training, the images are resized to $256\times256$ and we use irregular masks as in \cite{liu2018image}. We use Adam optimizer and follow \cite{Nazeri_2019_ICCV} to set $\beta_{1}=0$ and $\beta_{2}=0.9$. The learning rate = $2.92\times10^{-4}$ and $2.92\times10^{-5}$ for discriminator, with a learning rate decay ratio of $0.78$. Batch size = $4$.



\begin{table}[ht]
\caption{Valid Possibility on Cleft Lip dataset.}
\vspace{-1.6em}
\begin{center}

\begin{tabular}{|c|c|c|c|c|}
\hline
Method &EC\cite{Nazeri_2019_ICCV}&Lafin\cite{yang2019lafin}&CSTDG\cite{guo2021image}&Ours\T\B\\
\cline{1-5} 
CleftLip10& 0.233 & 0.233& 0.233&\textbf{0.5}\T\B\\
\cline{1-5}
CleftLip24&0.319 &0.222&0.264&\textbf{0.333}\T\B\\
\hline
\end{tabular}
\vspace{-2.6em}
\label{tab1}
\end{center}
\end{table}

\begin{table}[ht]
\caption{Average Ranking on the Cleft Lip dataset.}
\vspace{-1.6em}
\begin{center}

\begin{tabular}{|c|c|c|c|c|}
\hline
Method &EC\cite{Nazeri_2019_ICCV}&Lafin\cite{yang2019lafin}&CSTDG\cite{guo2021image}&Ours\T\B\\
\cline{1-5} 
CleftLip10& 1.857 & 1.714& 2.429&\textbf{1.267}\T\B\\
\cline{1-5}
CleftLip24&1.696 &1.813&1.947&\textbf{1.208}\T\B\\
\hline
\end{tabular}
\vspace{-1.6em}
\label{tab2}
\end{center}
\end{table}

\begin{figure}[htbp]
\vspace{-1em} 
\centerline{\includegraphics[scale=0.32]{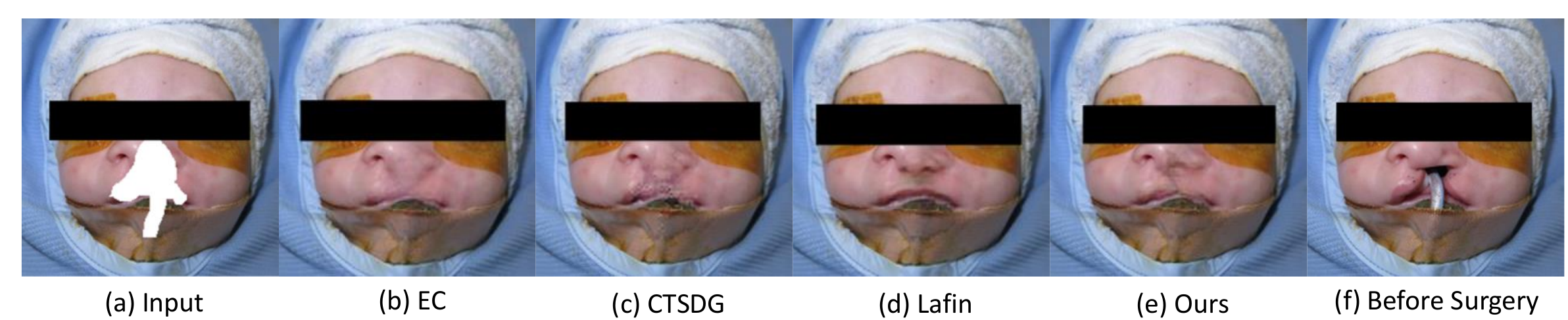}}
\vspace{-1em} 
\caption{Visual comparison of different facial inpainting methods on real Cleft Lip dataset: (a) input masked image, (b) EdgeConnect \cite{Nazeri_2019_ICCV}, (c) CTSDG \cite{guo2021image}, (d) Lafin \cite{yang2019lafin}, (e) Ours, and (f) Before Surgery}
\label{exmp}
\vspace{-1em} 
\end{figure}

\subsection{Experimental Validation}

\subsubsection{Cleft Lips Repair}
We use the CleftLip10 and CleftLip24 as test sets to compare our model with the current state-of-the-art facial inpainting methods \cite{Nazeri_2019_ICCV,yang2019lafin,guo2021image}. The visualization results are shown in Fig.~\ref{exmp}. We crop and resize them to $256\times256$, then design a mask to cover the cleft lip, as well as the medical equipment used during surgery, according to the type (unilateral and bilateral) and the severity of cleft lip for each patient. To better evaluate the feasibility of the proposed method, we invited NHS specialist cleft lip surgeons to assess the results based on the quality, consistency and validity. For each patient, results from four models are presented together. To avoid bias, the results are mixed and unlabelled. Images are deemed invalid if it is excessively blurry or illogical, e.g. flying lip or three nostril (see Fig.~\ref{exmp}(d)).
The valid probability represents the the success rate of models in repairing cleft lips images (see Table~\ref{tab1}), and the average ranking represents the performance of each models in the valid repaired results (see Table~\ref{tab2}).

\begin{table}[ht]
\caption{Quantitative Comparison on CelebA.}
\begin{center}

\begin{tabular}{c|c|c|c|c}
\hline
\hline
Mask Ratio&Model&PSNR&SSIM&FID\T\B\\

\hline
\hline
&EC\cite{Nazeri_2019_ICCV}&36.1340&0.9880&0.4706\T\\
0-20\%&Lafin\cite{yang2019lafin}&35.9544&0.9870&0.5845\\
&CTSDG\cite{guo2021image}&37.9275&0.9908&\textbf{0.3420}\\
&Ours&\textbf{38.1083}&\textbf{0.9911}&0.3421\B\\
\hline
&EC\cite{Nazeri_2019_ICCV}&28.3684 & 0.9486 & 3.1275\T\\
20-40\%&Lafin\cite{yang2019lafin}&28.2797&0.9476&3.3880\\
&CTSDG\cite{guo2021image}&29.3860&0.9570&2.8436\\
&Ours&\textbf{29.6678}&\textbf{0.9595}&\textbf{2.8327}\B\\
\hline
&EC\cite{Nazeri_2019_ICCV}&23.4513&0.8561&6.1253\T\\
40-60\%&Lafin\cite{yang2019lafin}&23.5109&0.8614&6.5367\\
&CTSDG\cite{guo2021image}&\textbf{24.3130}&\textbf{0.8762}&8.7051\\
&Ours&24.2076&0.8726&\textbf{4.3419}\B\\
\hline
\hline
\end{tabular}
\vspace{-2.6em} 
\label{tab3}
\end{center}
\end{table}

From our observation, each of the four models is capable for repairing small cleft lip areas. However, our method performs best for relatively complex situation, such as Fig.~\ref{exmp}(f) with severe cleft lips and large medical equipment.
The result from EC \cite{Nazeri_2019_ICCV} is too blurry and CTSDG \cite{guo2021image} leads obvious artifacts in regenerated region. 
Lafin \cite{yang2019lafin} seems to be suffering from model collapse and was seriously misled by the input indicator, generating a full nose at the right nostril.
From the surgeons assessment, our model generates more natural and semantically plausible images with a higher valid possibility.
Additionally, our model is able to generate textures similar to post-surgical scars while we leave certain intimation to the model (see Fig.~\ref{exmp}(e)).

\subsubsection{Facial Inpainting}
We compare our model with current state-of-the-art facial inpainting models on CelebA. The evaluation metrics involve peak signal-to-noise ratio (PSNR), structural similarity index (SSIM) \cite{wang2004image} and Frechet Inception Distance (FID) \cite{heusel2017gans}, which is shown in Table \ref{tab3}. Higher PSNR, SSIM, and lower FID, indicate better generated image quality. 
We observe that our model overall suppresses state-of-the-art inpainting models in terms of small and medium masked ratio. The latest CTSDG outperforms ours by a small margin in large missing regions case in terms of PSNR and SSIM, but it is much lower than ours in FID.

\begin{table}[ht]
\caption{Ablation Study on CelebA.}
\begin{center}

\begin{tabular}{c|c|c|c|c|c}
\hline
\hline
&Mask & \multicolumn{3}{|c|}{Irregular Mask} &Regular\T\B\\
\cline{2-5} 
&Mask Ratio& 0-20\% & 20-40\%& 40-60\%& Mask\T\B\\
\hline
\hline
&Baseline &37.1742 &29.1189 &23.7541 &25.9412\T\\
PSNR&Base+Lmk&37.3932&29.2115&23.7948&26.074\\
&Ours&\textbf{38.1083}&\textbf{29.6678}&\textbf{24.2076}&\textbf{26.685}\B\\
\hline
&Baseline&0.9895&0.9545&0.8605&0.9113\T\\
SSIM&Base+Lmk&0.9897&0.9554&0.8626&0.9144\\
&Ours&\textbf{0.9911}&\textbf{0.9595}&\textbf{0.8726}&\textbf{0.9231}\B\\
\hline
&Baseline&0.5330&3.3655&5.7635&3.6037\T\\
FID&Base+Lmk&0.4614&2.9762&5.1124&3.410\\
&Ours&\textbf{0.3421}&\textbf{2.8327}&\textbf{4.3419}&\textbf{3.274}\B\\
\hline
\hline
\end{tabular}
\vspace{-1.6em} 
\label{tab4}
\end{center}
\end{table}

To validate the effectiveness of our multi-task architecture, we remove the parameter sharing between two tasks and take encoder followed by image generator as the baseline.
Then, we implement the landmark predictor (Base+Lmk) and gated convolution (Ours) progressively. 
As shown in Table ~\ref{tab4}, the integration of both the multi-task model and gated convolutions improve the performance on both irregular and regular masks.

\section{Conclusion}
In this paper, we propose a novel approach to provide an guidance image for cleft lip surgery by masking the cleft lip part and generating lip and nose without cleft. To achieve this task, we design a multi-task image inpainting model that can better protect patient privacy. We collected two real-world patient datasets to demonstrate the feasibility of proposed approach. Three expert cleft lip surgeons assessed that our design outperforms state-of-the-art methods in both valid possibility and image quality, while the performance of our model on CelebA also suppresses the state-of-the-art facial inpainting counterparts.
\section{Acknowledgment}
We would like to acknowledge Ross Obukofe for his support in this project.

\bibliographystyle{IEEEtran}
\bibliography{mybib}

\end{document}